\documentclass[letterpaper, 10 pt, conference]{ieeeconf}  

\IEEEoverridecommandlockouts                              

\usepackage{amsmath} 
\usepackage{amssymb}  
\usepackage{mathtools}
\usepackage{xcolor}

\usepackage{eqnarray}
\usepackage{graphicx}
\usepackage{multirow}
\usepackage{tabularx}
\graphicspath{{images}}
\usepackage{xcolor}

\usepackage{comment}
\usepackage{cite}

\usepackage[font=footnotesize]{subcaption}
\usepackage[font=footnotesize]{caption}

\usepackage{threeparttable}
\usepackage{booktabs}

\newcommand{\pa}{HAMLET}
\newcommand{\hattn}{MAT}
\newcommand{\fpa}{HAMLET: \textbf{H}ierarchic\textbf{a}l \textbf{M}ultimoda\textbf{l} S\textbf{e}lf-a\textbf{t}tention based HAR}
\newcommand{\fhattn}{MAT: \textbf{M}ultimodal \textbf{A}ten\textbf{t}ion based Feature Fusion}
\newcommand{\uat}{UAT}

\title{\LARGE \bf
HAMLET: A Hierarchical Multimodal Attention-based Human Activity Recognition Algorithm
}

\author{Md Mofijul Islam$^{1}$ and Tariq Iqbal$^{1}$
\thanks{$^{1}$The authors are with the Dept. of Engineering Systems and Environment, Univ. of Virginia, USA. {\tt\small \{mi8uu,tiqbal\}@virginia.edu}.}   
}

\begin{document}

\maketitle
\thispagestyle{empty}
\pagestyle{empty}

\begin{abstract}
To fluently collaborate with people, robots need the ability to recognize human activities accurately. Although modern robots are equipped with various sensors, robust human activity recognition (HAR) still remains a challenging task for robots due to difficulties related to multimodal data fusion. To address these challenges, in this work, we introduce a deep neural network-based multimodal HAR algorithm, {\pa }. {\pa } incorporates a hierarchical architecture, where the lower layer encodes spatio-temporal features from unimodal data by adopting a multi-head self-attention mechanism. We develop a novel multimodal attention mechanism for disentangling and fusing the salient unimodal features to compute the multimodal features in the upper layer. Finally, multimodal features are used in a fully connect neural-network to recognize human activities. We evaluated our algorithm by comparing its performance to several state-of-the-art activity recognition algorithms on three human activity datasets. The results suggest that {\pa } outperformed all other evaluated baselines across all datasets and metrics tested, with the highest top-1 accuracy of 95.12\% and 97.45\% on the UTD-MHAD \cite{utd_mhad} and the UT-Kinect\cite{ut_kinect} datasets respectively, and F1-score of 81.52\% on the UCSD-MIT \cite{mit_ucsd} dataset. We further visualize the unimodal and multimodal attention maps, which provide us with a tool to  interpret the impact of attention mechanisms concerning HAR. 
\end{abstract}
\section{Introduction}
\label{sec:introduction}

Robots are sharing physical spaces with humans in various collaborative environments, from manufacturing to assisted living to healthcare \cite{Riek2017HealthCare,iqbal2019human,Iqbal2016T-RO}, to improve productivity and to reduce human cognitive and physical workload \cite{andi_iros}. To be effective in close proximity to people, collaborative robotic systems (CRS) need the ability to automatically and accurately recognize human activities \cite{fosapt}. This capability will enable CRS to operate safely and autonomously to work alongside human teammates \cite{iqbal2017coordination}.

To fluently and fluidly collaborate with people, CRS needs to recognize the activities performed by their human teammates robustly \cite{tiqbal_joint_action,Iqbal2015TAC,mit_ucsd}. Although modern robots are equipped with various sensors, robust human activity recognition (HAR) remains a fundamental problem for CRS \cite{iqbal2019human}. This is partly because fusing multimodal sensor data efficiently for HAR is challenging. 
Therefore, to date, many researchers have focused on recognizing human activities by leveraging on a single modality, such as visual, pose or wearable sensors \cite{andi_iros,new_sk_rep,st_graph_sk,space_time_sk_review,iqbal2016tempo}. However, HAR models reliant on unimodal data often suffer a single point feature representation failure. For example, visual occlusion, poor lighting, shadows, or complex background can adversely affect only visual sensor-based HAR methods. Similarly, noisy data from accelerometer or gyroscope sensors can reduce the performance of HAR methods solely depending on these sensors \cite{mit_ucsd,multimodal_survey}.

\begin{figure}[!t]
    \centering
    \includegraphics[width=0.9\columnwidth]{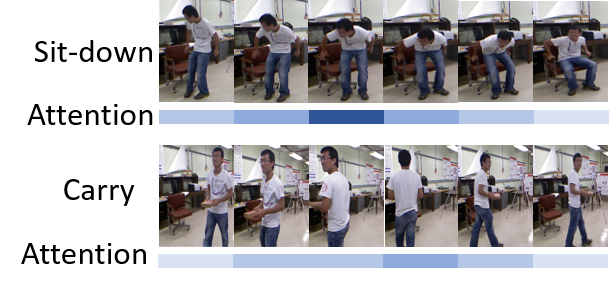}
    \caption{Example of two activities (\textit{Sit-Down} and \textit{Carry}) from the UT-Kinect dataset (the first row). The second row presents the temporal-attention weights on the corresponding RGB frames using {\pa }. For these sequences, {\pa } pays more attention to the third RGB image segment for the \textit{Sit-Down} activity (top) and on the fourth RGB image segment for the \textit{Carry} activity (bottom). Here, a lighter color represents a lower attention.}
    \label{fig:hat}
     \vspace{-0.3in}
\end{figure}

Several approaches have been proposed to overcome the weaknesses of the unimodal methods by fusing multimodal sensor data that can provide complementary strengths to achieve a robust HAR  \cite{multimodal_survey,mit_ucsd,self_attention_iros19,keyless,fusion_approaches,Hasan_2019}. 
Although many of these approaches exhibit robust performances than unimodal HAR approaches, there remain several challenges that prevent these methods from efficiently working on CRSs \cite{multimodal_survey}. 
For example, while fusing data from multiple modalities, these methods rely on a fixed-fusion approach, e.g., concatenate, average, or sum. Although one type of fusion approach works for a specific activity, these approaches can not provide any guaranty that the same performance can be achieved on a different activity class using the same merging method. Moreover, these proposed approaches provide uniform weightage on the data from all modalities. However, depending on the environment, one sensor modality may provide more enhanced information than the other sensor modality. For example, a visual sensor may provide valuable information about a gross human activity than a gyroscope sensor data, which a robot needs to learn from data automatically. Thus, these approaches can not provide robust HAR for CRSs.

To address these challenges, in this work, we introduce a novel multimodal human activity recognition algorithm, called {\fpa } algorithm for CRS. 
{\pa } first extracts the spatio-temporal salient features from the unimodal data for each modality. {\pa } then employs a novel multimodal attention mechanism, called \fhattn, for disentangling and fusing the unimodal features. These fused multimodal features enable {\pa } to achieve higher HAR accuracies (see Sec.~\ref{sec:proposedApproach}).

The modular approach to extract spatial-temporal salient features from unimodal data allows {\pa } to incorporate pre-trained feature encoders for some modalities, such as pre-trained ImageNet models for RGB and depth modalities. This flexibility enables {\pa} to incorporate deep neural network-based transfer learning approaches. Additionally, the proposed novel multimodal fusion approach (MAT) utilizes a multi-head self-attention mechanism, which allows {\pa } to be robust in learning weights of different modalities based on their relative importance in~HAR from data.

We evaluated {\pa } by assessing its performance on three human activity datasets (UCSD-MIT\cite{mit_ucsd}, UTD-MHAD\cite{utd_mhad} and UT-Kinect \cite{ut_kinect}) compared with several state-of-the-art activity recognition algorithms from prior literature (\cite{keyless,mit_ucsd,posemap,sdd_iccv,mcrl,sos,jdm_cnn,dcnn,dmm_mff,utd_mhad}) and two baseline methods (see  Sec.~\ref{sec:experimenResults}).
In our empirical evaluation, {\pa } outperformed all other evaluated baselines across all datasets and metrics tested, with the highest top-1 accuracy of 95.12\% and 97.45\% on the UTD-MHAD \cite{utd_mhad} and the UT-Kinect\cite{ut_kinect} datasets respectively, and F1-score of 81.52\% on the UCSD-MIT \cite{mit_ucsd} dataset (see Sec.~\ref{sec:results_and_discussion}).
We visualize an attention map representing how the unimodal and the multimodal attention mechanism impacts multimodal feature fusion for HAR (see Sec.~\ref{sec:attention_vis}).

\section{Related Works}
\label{sec:relatedWorks}

\textbf{Unimodal HAR:} Human activity recognition has been extensively studied by analyzing and employing the unimodal sensor data, such as skeleton, wearable sensors, and visual (RGB or depth) modalities \cite{hussein2013human}.  As generating hand-crafted features is found to be a difficult task, and these features are often highly domain-specific, many researchers are now utilizing the deep neural network-based approaches for human activity recognition.

Deep learning-based feature representation architectures, especially convolutional neural networks (CNNs) and long-short-term memory (LSTM), have been widely adopted to encode the spatio-temporal features from visual (i.e., RGB and depth) \cite{new_sk_rep,co_occurrence,closer_look_sp,sp_temporal_relation,sp_3d_conv,slowfast} and non-visual (i.e., sEMG and IMUs) sensors data \cite{andi_iros,mit_ucsd,totty2017muscle}. For example, Li et al. \cite{co_occurrence} developed a CNN-based learning method to capture the spatio-temporal co-occurrences of skeletal joints. To recognizing human activities from video data, Wang et al. proposed a 3D-CNN and LSTM-based hybrid model to detect compute salient features \cite{sp_3d_conv_lstm}. 
Recently, the graphical convolutional network has been adopted to find spatial-temporal patterns in unimodal data \cite{st_graph_sk}.


Although these deep-learning-based HAR methods have shown promising performances in many cases, these approaches rely significantly on modality-specific feature embeddings. If such an encoder fails to encode the feature properly because of noisy data (e.g., visual occlusion or missing or low-quality sensor data), then these activity recognition methods suffer to perform correctly.




\textbf{Multimodal HAR:} Many researchers have started working on designing multimodal learning methods by utilizing the complementary features from different modalities effectively to overcome the dependencies on a single modality data of modality-specific HAR models \cite{Garcia_2018_ECCV,keyless,self_attention_iros19,joze2019mmtm}. One crucial challenge that remains in developing a multimodal learning model is to fuse the various unimodal features efficiently.

Several approaches have been proposed to fuse data from similar modalities \cite{two_stream_cnn,sp_two_stream_residual,sp_mul_motion_gating,conv_fusion,zhang2018fusing}. For example, Simonyan et al. proposed a two-stream CNN-based architecture, where they incorporated a spatial CNN network to capture the spatial features, and another CNN-based temporal network to learn the temporal features from visual data \cite{two_stream_cnn}.  As CNN-based two-stream network architecture allows to appropriately combine the spatio-temporal features, it has been studied in several recent works, e.g., residual connection in streams \cite{sp_two_stream_residual}, convolutional fusion \cite{conv_fusion} and slow-fast network~\cite{slowfast}.

Other works have focused on fusing features from various modalities, i.e., fusing features from visual (RGB), pose, and wearable sensor modalities simultaneously \cite{multimodal_survey,mfas,joze2019mmtm}. M\"{u}nzner et al. \cite{fusion_approaches} studied four types of feature fusion approaches: early fusion, sensor and channel-based late fusion, and shared filters hybrid fusion. They found that the late and hybrid fusion outperformed early fusion. Other approaches have focused on fusing modality-specific features at a different level of a neural network architecture \cite{mfas}. For example, Joze et al. \cite{joze2019mmtm} designed an incremental feature fusion method, where the features are merged at different levels of the architecture.
Although these approaches have been proposed in the literature, generating multimodal features by dynamically selecting the unimodal features is still an open challenge.

\textbf{Attention mechanism for HAR:}
Attention mechanism has been adopted in various learning architectures to improve the feature representation as it allows the feature encoder to focus on specific parts of the representation while extracting the salient features \cite{attention,attention_effective_approach,xu2015show,lu2017knowing,keyless,mnih2014recurrent,lu2019vilbert,gao2019multi}. Recently, several multi-head self-attention based methods have been proposed, which permit to disentangle the feature embedding into multiple features (multi-head) and to fuse the salient features to produce a robust feature embedding \cite{transformer}. 

Many researchers have started adopting the attention mechanism in human activity recognition \cite{self_attention_iros19,keyless}. For example, Xiang et al. proposed a multimodal video classification network, where they utilized an attention-based spatio-temporal feature encoder to infer modality-specific feature representation \cite{keyless}. The authors explored the different types of multimodal feature fusion approaches (feature concatenation, LSTM fusion, attention fusion, and probabilistic fusion), and found that the concatenated features showed the best performance among the other fusion methods. To date, most of the HAR approaches have utilized attention-based methods for encoding the unimodal features. However, the attention mechanism has not been used for extracting and fusing salient features from multiple modalities.

To address these challenges, in our  proposed multimodal HAR algorithm (\pa),
we have designed a modular way to encode unimodal spatio-temporal features by adopting a multi-head self-attention approach. 
Additionally, we have developed a novel multimodal attention mechanism, {\hattn}, for disentangling and fusing the salient unimodal features to compute the multimodal features. 

\section{Proposed Modular Learning Method}
\label{sec:proposedApproach}

In this section, we present our proposed multimodal human-activity recognition method, called \fpa. We present the overall architecture in Fig.~\ref{fig:hat}. In {\pa }, the multimodal features are encoded into two steps, and those features are then used for activity recognition as follows:
\begin{itemize}
    \item At first, the Unimodal Feature Encoder module encodes the spatial-temporal features for each modality by employing a modality-specific feature encoder and a multi-head self-attention mechanism (UAT).
    \item In the second step, the Multimodal Feature Fusion module (MAT) fuses the extracted unimodal features by applying our proposed novel multimodal self-attention method.
    \item These computed multimodal features are then utilized by a fully connected neural network to calculate the probability of each activity class.
\end{itemize}


\begin{figure}[!t]
    \centering
    \includegraphics[width=\columnwidth]{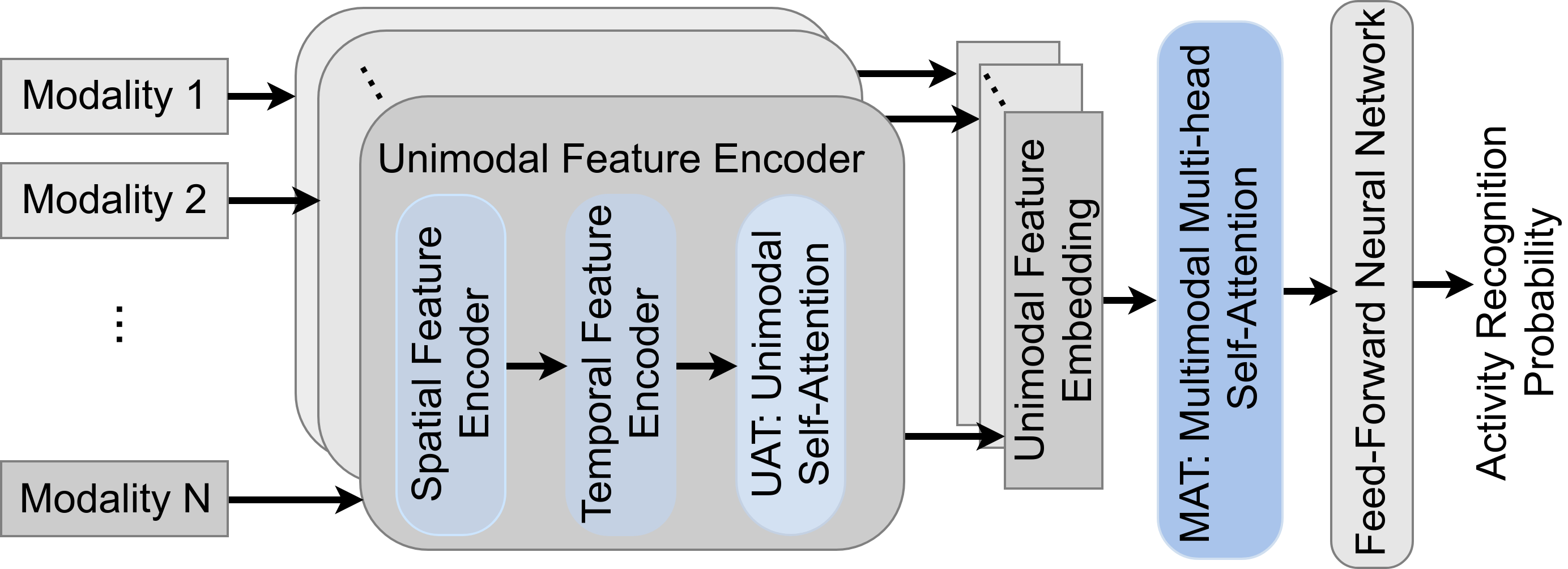}
    \caption{{\pa }: Hierarchical Multimodal Self-Attention based HAR.}
    \label{fig:hat}
    \vspace{-0.2in}
\end{figure}

\subsection{Unimodal Feature Encoder}
The first step of {\pa} is to compute a feature representation for data from every modality. To achieve that, we have designed modality-specific feature encoders to encode data from different modalities. The main reasoning behind this type of modality-specific modular feature encoder architecture is threefold. First, each of the modalities has different feature distribution and thus needs to have a different feature encoder architecture. For example, the distribution and representation of visual data differ from the skeleton and inertial sensor data. Second, the modular architecture allows incorporating unimodal feature encoders without interrupting the performance of the encoders of other modalities. This capability enables the modality-specific transfer learning. Thus we can employ a pre-trained feature encoder to produce robust feature representation for each modality. Third, the unimodal feature encoders can be trained and executed in parallel, which reduces the computation time during the training and inference phases.


Each of the unimodal feature encoders is divided into three separate sequential sub-modules: spatial feature encoder, temporal feature encoder, and unimodal attention module (UAT). Before applying a spatial feature encoder, at first the whole sequence of data $D^m=(d_1^m,d_2^m,...,d_T^m)$ from modality $m$ is converted into segmented sequence $X^m=(x_1^m,x_2^m,...,x_{S^m}^m)$ of size $B\times S^m \times E^m$, where $B$ is the batch size, $S^m$ and $E^m$ are the number of segments and feature dimension for modality $m$ respectively. In this work, we represent the feature dimension $E^m$ for RGB and depth modality as $(channel(C^m) \times height(H^m) \times width(W^m))$, where $C^m$ is the number of channels in an image. 

\subsubsection{Spatial Feature Encoder}
We used a temporal pooling method to encode segment-level features instead of extracting the frame-level features, similar to \cite{keyless}. We have implemented the temporal pooling for two reasons: first, as the successive frames represent similar features, it is redundant to apply spatial feature encoder on each frame, which increases the training and testing time. By Utilizing the temporal pooling, {\pa} reduces its computational time. Moreover, this polling approach is necessary to implement {\pa} on a real-time robotic system. Second, the application of recurrent neural networks for each frame is computationally expensive for a long sequence of data. We used adaptive temporal max-pool to pool the encoded segment level features.

As our proposed modular architecture allows modality-specific transfer learning, we have incorporated the available state-of-the-art pre-trained unimodal feature encoders. For example, we have incorporated ResNet50 to encode the RGB modality. We extend the convolutional co-occurrence feature learning method \cite{co_occurrence} to hierarchically encode segmented skeleton and inertial sensor data. In this work, we used two stacked 2D-CNNs architecture to encode co-occurrence features: first 2D-CNN encodes the intra-frame point-level information and second 2D-CNN extract the inter-frame features in a segment. Finally, spatial feature encoder for modality $m$ produces a spatial feature representation $F^{S}_{m}$ of size $(B\times S^m \times E^{S,m})$ from segmented $X^m$, where $E^{S,m}$ is the spatial feature embedding dimension. 

\subsubsection{Temporal Feature Encoder}
After encoding the segment level unimodal features, we employ recurrent neural networks, specifically unidirectional LSTM, to extract the temporal feature features $H^m=(h_1^m,h_2^m,...,h_s^m)$ of size $(B\times S^m \times E^{H,m})$ from $F^{S}_{m}$, where $E^{H,m}$ is the LSTM hidden feature dimension. Our choice of unidirectional LSTM over other recurrent neural network architectures (such as gated recurrent units) was based on the ability of LSTM units to capture long-term temporal relationships among the features. Besides, we need our model to detect human activities in real-time, which motivated our choice of unidirectional LSTMs over bi-directional LSTMs. 


\subsubsection{Unimodal Self-Attention ({\uat }) Mechanism}
\label{sec:msa}
The spatial and temporal feature encoder sequentially encodes the long-range features. However, it cannot extract salient features by employing sparse attention to the different parts of the spatial-temporal feature sequence. Self-attention allows the feature encoder to pay attention to the sequential features sparsely and thus produce a robust unimodal feature encoding. Taking inspiration from the Transformer-based multi-head self-attention methods \cite{transformer}, {\uat } combines the temporal sequential salient features for each modality. As each modality has its unique feature representation, the multi-head self-attention enables the {\uat } to disentangle and attend salient unimodal features. 

To compute the attended modality-specific feature embedding $F^{a}_{m}$ for modality $m$ using unimodal multi-head self-attention method, at first we need to linearly project the spatial-temporal hidden feature embedding $H^m$ to create query ($Q^m_i$), key ($K^m_i$) and value ($V^m_i$) for head $i$ in the following way,
\begin{eqnarray}
    Q_i^m &=& H^m W_i^{Q,m}\\
    K_i^m &=& H^m W_i^{K,m}\\
    V_i^m &=& H^m W_i^{V,m}
\end{eqnarray}
Here, each modality $m$ has its own projection parameters, $W_i^{Q,m} \in \mathbb{R}^{E^{H,m}\times E^K}, W_i^{K,m} \in \mathbb{R}^{E^{H,m}\times E^K}$, and $W_i^{V,m} \in \mathbb{R}^{E^{H,m}\times E^V}$, where $E^K$ and $E^V$ are projection dimensions, $E^K=E^V=E^{H,m}/h^m$, and $h$ is the total number of heads for modality $m$. After that we used scaled dot-product softmax approach to compute the attention score for head $i$ as:
\begin{eqnarray}
    Attn(Q_i^m, K_i^m, V_i^m) &=& \sigma \left( \frac{Q_i^m K_i^{m^T}}{\sqrt{d_k^m}} \right)V_i^m\\
    head_i^m &=& Attn(Q_i^m, K_i^m, V_i^m)
\end{eqnarray}
After that, all the head feature representation is concatenated and projected to produce the attended feature representation, $F^{a}_{m}$ in the following way, 
\begin{eqnarray}
    F^{a}_{m} = [head_1^m;...;head_h^m]W^{O,m}
\end{eqnarray}
Here, $W^{O,m}$ is the projection parameters of size $E^{H,m} \times E^{H}$, and the shape of $F^{a}_{m}$ is $(B\times S^m \times E^H)$, where $E^H$ is the attended feature embedding size. We used the same feature embedding size $E^H$ for all modalities to simplify the application of multimodal attention {\hattn } for fusing all the modality-specific feature representation, which is presented in the next section~\ref{sec:mat}. However, our proposed multimodal attention based feature fusion method can handle different unimodal feature dimensions. Finally, we fused the attended segmented sequential feature representation $F^{a}_{m}$ to produce the local unimodal feature representation $F_{m}$ of size $(B \times E^H)$. We can use different types of fusion to combine the spatio-temporal segmented feature encodings, such as sum, max, or concatenation. However, the concatenation fusion method is not a suitable approach to fuse large sequences, whereas max fusion may lose the temporal feature embedding information. As the sequential feature representations produced from the same modality, we have used the sum fusion approach to fuse attended unimodal spatial-temporal feature embedding $F^{a}_{m}$,
\begin{eqnarray}
    F_{m} = \sum_{s \in S^m} F^{a}_{m,s}
\end{eqnarray}

\begin{figure}[!t]
    \centering
    \includegraphics[width=\columnwidth]{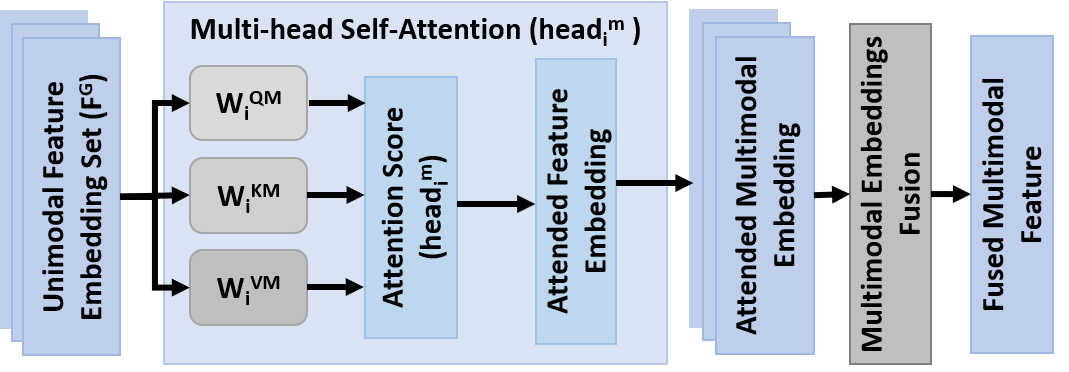}
    \caption{{\hattn :} Multimodal Attention-based Feature Fusion Architecture.}
    \label{fig:mma}
     \vspace{-0.2in}
\end{figure}

\subsection{Multimodal Feature Fusion}
\label{sec:mat}

In this work, we developed a novel multimodal feature fusion architecture based on our proposed multi-head self-attention model, \fhattn, which is depicted in Fig.~\ref{fig:mma}. After encoding the unimodal features using the modular feature encoders, we combine these feature embeddings $F_{m}$ in an unordered multimodal feature embedding set $F^{G^u}=(F_1, F_2,..., F_{M})$ of size $(B\times M \times D^{H})$, where $M$ is the total number of modalities. After that, we fed the set of unimodal feature representations $F^{G^u}$ into \hattn, which produces the attended fused multimodal feature representation $F^{G^a}$.

The multimodal multi-head self-attention computation is almost similar to the self-attention method described in Section~\ref{sec:msa}. However, there are two key differences. First, unlike encoding the positional information using LSTM to produce the sequential spatial-temporal feature embedding before applying the multi-head self-attention, in {\hattn}, we combine all the modalities feature embeddings without encoding any positional information. Also, {\hattn } and {\uat } modules have separate multi-head self-attention parameters. Second, after applying the multimodal attention method on the extracted unimodal features, we used two fusion approaches to fused the multimodal features:

\begin{itemize}
    \item MAT-SUM: extracted unimodal features are summed after applying the multimodal attention
    \begin{eqnarray}
       F^{G} &=& \sum_{m=1}^{M} F^{G^a}_m
    \end{eqnarray}
    \item MAT-CONCAT: in this approach the attended multimodal features are concatenated
    \begin{eqnarray}
F^{G} &=& [F^{G^a}_1;F^{G^a}_2;...;F^{G^a}_{M}]
    \end{eqnarray}
\end{itemize}

\subsection{Activity Recognition}
Finally, the fused multimodal feature representation $F^G$ is passed through a couple of fully-connected layers to compute the probability for each activity class. For aiding the learning process, we applied activation, dropout, batch normalization in different parts of the learning architecture (see the section~\ref{sec:implementation_details} for the implementation details).
As all the tasks of human-activity recognition, which we addressed in this work, are multiclass classification, we trained the model using cross-entropy loss function, mini-batch stochastic gradient optimization with weight decay regularization \cite{adamw}.
\begin{equation}
    loss(y,\hat{y}) = \frac{1}{B} \sum_{i=1}^{B} y_i \log{\hat{y_i}}
\end{equation}
\section{Experimental Setup}
\label{sec:experimenResults}

\begin{table}[t]
\centering
\caption{Performance comparison (mean top-1 accuracy) of multimodal fusion methods in {\pa} on UT-Kinect dataset \cite{ut_kinect}}
    \label{tab:com_fusion_on_ut_kinect}
\begin{tabular}{ccccccc}
\toprule
    \multicolumn{2}{c}{Number of Heads} & \multicolumn{2}{c}{Fusion Method} \\ \hline
    UAT & MAT & MAT-SUM & MAT-CONCAT \\ \hline 
    1 & 1 & 87.97 & 88.50 \\ 
    \textbf{1} & \textbf{2} & \textbf{93.50} & \textbf{97.45} \\ 
    2 & 2 & 92.50 & 93.00 \\ 
    2 & 4 & 93.50 & 94.50 \\
\bottomrule
\end{tabular}
\label{tab:accuracy33}
    \vspace{-0.2in}
\end{table}


\subsection{Datasets}
We evaluated the performance of our proposed multimodal HAR method, \pa, using three human-activity datasets:  UTD-MHAD \cite{utd_mhad}, UT-Kinect \cite{ut_kinect}, UCSD-MIT \cite{mit_ucsd}.

\par \textbf{UTD-MHAD} \cite{utd_mhad} human activity dataset consists of a total of 27 human actions covering from sports, to hand gestures, to training exercises and daily activities. Eight people repeated each action for four times. After removing the corrupted sequences, this dataset contains a total of 861 data samples. 

\textbf{UT-Kinect} \cite{ut_kinect} dataset contains a total of ten indoor daily life activities (e.g., walking, standing up, etc.) with three modalities: RGB, depth, and 3D skeleton. Each activity was performed two times by each person. Thus there were a total of 200 activity samples in this dataset.

\textbf{UCSD-MIT} \cite{mit_ucsd} human activity dataset consists of eleven sequential activities in an automotive assembly task. Each assembly task was performed five people, and each person performed the task for five times. This dataset contains there modalities: 3D skeleton data from a motion capture system, and sEMG and IMUs data from a wearable sensor.

\subsection{Implementation Details}
\label{sec:implementation_details}
\textbf{Spatial-temporal feature encoder:} We incorporated pre-trained ResNet50 for encoding the RGB and depth data \cite{resnet}. We applied max pooling with a kernel size of five and stride of three for pooling segment level features. We extended the co-occurrence \cite{co_occurrence} feature extraction network to encode segmented skeleton and inertial sensor features. Finally, for capturing the temporal features, we used a two-layer unidirectional LSTM. We used embedding size 128 and 256 for UCSD-MIT\cite{mit_ucsd} and UT-Kinect\cite{ut_kinect} spatial-temporal features embedding respectively.

\par \textbf{Hyper-parameters and optimizer:} We utilized the pre-trained ResNet architecture for encoding RGB and depth modality. However, in the case of a co-occurrence feature encoder (skeleton and inertial sensor), we applied BatchNorm-2D, ReLu activation, and Dropout layers sequentially. After encoding each unimodal features, we applied ReLu activation and Dropout. Finally, in {\hattn}, after fusing the multimodal features, we used BatchNorm-1D, ReLu activation, and Dropout sequentially. We varied the dropout probability between $0.2-0.4$ in different layers. In multi-head self-attention for both unimodal and multimodal feature encoders, we varied the number of heads from one to eight. 
We train the learning model using Adam optimizer with weight decay regularization option \cite{adamw} and cosine annealing warm restarts \cite{warm_restart} with an initial learning rate set to $3e^{-4}$.


\par \textbf{Training environment:}
We implemented all the parts of the learning model using Pytorch-1.4 deep learning framework \cite{paszke2019pytorch}. We trained our model in different types of GPU-based computing environments (GPUs: P100, V100, K80, and RTX6000).

\subsection{State-of-the-art Methods and Baselines}
\label{sec:baselines}
We designed two baseline HAR methods and reproduce a state-of-art HAR method to evaluate the impact of attention method in encoding and fusing multimodal features:
\begin{itemize}
    \item \textbf{Baseline-1 (NSA)} does not use the attention mechanism for encoding unimodal or fusing multimodal features. 
    \item \textbf{Baseline-2 (USA)} only applies multi-head self-attention to encode unimodal features but fuses the multimodal embedding without applying attention. This baseline method is similar to the self-attention based multimodal HAR proposed in \cite{self_attention_iros19}.
    \item \textbf{Keyless Attention} \cite{keyless} employed an attention mechanism to encode the modality-specific features. However, it did not utilize attention methods to fuse the multimodal features, instead those were concatenated.
\end{itemize}

\subsection{Evaluation metrics}
To evaluate the accuracy of {\pa}, the Keyless Attention model \cite{keyless}, the NSA, and the USA algorithms, we performed leave-one-actor-out cross-validation across
all the trials for each person on each dataset. Similar to the original evaluation schemes, we reported activity recognition accuracy for the UT-Kinect \cite{ut_kinect} and the UTD-MHAD datasets \cite{utd_mhad}, and F1-score (in \%) for the UCSD-MIT dataset \cite{mit_ucsd}.

\par  
To evaluate {\pa }, the Keyless attention method, and baseline methods on UT-Kinect and UTD-MHAD datasets, we used RGB and skeleton data. We leveraged skeleton, IMUs, and sEMG modalities on the UCSD-MIT dataset. 

\section{Results and Discussion}
\label{sec:results_and_discussion}

\begin{table}[!t]
    \centering
    \caption{Performance comparison (mean top-1 accuracy) of multimodal HAR methods on UT-Kinect dataset \cite{ut_kinect}}
    \label{tab:com_on_ut_kinect}
    \begin{tabular}{llc}
        \toprule
        \multirow{1}{*}{Method}& {Fusion Type} & Top-1 Accuracy (\%) \\
        \hline
        \multirow{2}{*}{NSA} & SUM & 54.34 \\
        & CONCAT & 52.31\\
        \hline
        \multirow{2}{*}{USA} & SUM & 55.82 \\
        & CONCAT & 54.34 \\
        \hline
        \multirow{1}{*}{{KEYLESS \cite{keyless} (2018)}} & CONCAT & 94.50\\
        \hline
        \multirow{2}{*}{\textbf{\pa}} & MAT-SUM & 95.56\\
        & \textbf{MAT-CONCAT} & \textbf{97.45}\\
        \bottomrule
    \end{tabular}
    \vspace{-0.2in}
\end{table}


\subsection{Multimodal Attention-based Fusion Approaches}
\label{sec:attention_fusion_approach}

We first evaluated the accuracy of two multimodal attention-based feature fusion approaches of \pa: MAT-SUM and MAT-CONCAT. We also varied the number of heads used in UAT and MAT steps to determine the optimal configuration of these values.

\par \textbf{Results:} We evaluated UAT and MAT attention methods as well as the fusion approaches (MAT-SUM and MAT-CONCAT) on the UT-Kinect dataset \cite{ut_kinect}, presented in Table~\ref{tab:com_fusion_on_ut_kinect}. We used the RGB and skeleton modalities and reported top-1 accuracy by following the original evaluation scheme. The results suggest that the MAT-CONCAT fusion method showed the highest top-1 accuracy (97.45\%), with one and two heads in UAT and MAT methods, respectively.

\par \textbf{Discussion:} The results suggest the concatenation-based fusion approach (MAT-CONCAT) performed better than the summation-based fusion approach (MAT-SUM). Because the MAT-CONCAT allows {\hattn } to disentangle and apply attention mechanisms on the unimodal features to generate robust multimodal features for activity classification. On the other hand, the sum-based fusion method merged the unimodal features into a single representation, which makes it difficult for {\hattn } to disentangle and apply appropriate attention to unimodal features.

\par The results from Table~\ref{tab:com_fusion_on_ut_kinect} also indicate an improvement in activity recognition accuracy with the increment of the number of heads in the MAT when keeping the number of heads fixed in the UAT. However, this relationship does not hold when the number of heads was changed in the UAT. As a large number of heads reduce the size of feature embedding, increasing the number of heads in the UAT may result in an inadequate feature representation. Thus, based on the size of the features used in this work, the results suggest that one head in the UAT and two heads in the MAT methods display the best accuracy. Thus, we utilized these values for further evaluations. 

\begin{table}[!t]
\centering
\caption{Performance comparison (mean top-1 accuracy) of multimodal fusion methods on UTD-MHAD dataset \cite{utd_mhad}}
    \label{tab:com_utd_mhad}
\begin{tabular}{ccc}
\toprule
    Method & Year & Top-1 Accuracy (\%) \\ \hline
    Kinect \& Inertial \cite{utd_mhad} & 2015 & 79.10 \\
    DMM-MFF \cite{dmm_mff} & 2015 & 88.40 \\
    DCNN \cite{dcnn} & 2016 & 91.2\\
    JDM-CNN \cite{jdm_cnn} & 2017 & 88.10 \\
    S$^2$DDI \cite{sdd_iccv} & 2017 & 89.04 \\
    SOS \cite{sos} & 2018 & 86.97 \\
    MCRL \cite{mcrl} & 2018 & 93.02 \\
    PoseMap \cite{posemap} & 2018 & 94.51 \\ 
    \textbf{\pa{ }(MAT-CONCAT)} & - & \textbf{95.12} \\
\bottomrule
\end{tabular}
\label{tab:accuracy33}
    \vspace{-0.2in}
\end{table}

\subsection{Comparison with Multimodal HAR Methods}\label{sec:com_multimodal_har}
As {\pa } takes a multimodal approach, it is reasonable to evaluate the accuracy against the state-of-the-art multimodal approaches. Thus, we compare the performance of {\pa } with two baseline methods (the USA and the NSA, see Sec.~\ref{sec:baselines}) and several state-of-the-art multimodal approaches. We presented the results in Tables~\ref{tab:com_on_ut_kinect} (UT-Kinect),~\ref{tab:com_utd_mhad} (UTD-MHAD) \&~\ref{tab:com_on_mit_ucsd} (UCSD-MIT).


{\textbf{Results:}} In the UT-Kinect dataset, RGB and skeleton modalities have been used to train the learning models. Following the original evaluation scheme, we report the top-1 accuracy in Table~\ref{tab:com_on_ut_kinect}. The results indicate that {\pa } achieved the highest 97.45\% top-1 accuracy across all other methods.

We also evaluate the performance of {\pa } on the UTD-MHAD \cite{utd_mhad} dataset. We train and test {\pa } on RGB and Skeleton data and report the top-1 accuracy while using MAT-CONCAT in Table~\ref{tab:com_utd_mhad}. The results suggest that {\pa } outperformed all the evaluated state-of-the-art baselines and achieved the highest accuracy of 95.12\%. 

\par For the UCSD-MIT dataset, all the learning methods are trained on the skeleton, inertial, and sEMG data. All the training models have been used late or intermediate fusion except for the results presented from \cite{mit_ucsd}, which used an early feature fusion approach. In Table~\ref{tab:com_on_mit_ucsd}, the results suggest that {\pa } with MAT-SUM fusion method outperformed the baselines and state-of-the-art works by achieving the highest 81.52\% F1-score (in \%).  

\begin{table}[!t]
    \centering
    \caption{Performance comparison (mean F1-scores in \%) of multimodal HAR methods on UCSD-MIT dataset \cite{mit_ucsd}}
    \label{tab:com_on_mit_ucsd}
    \begin{tabular}{llc}
        \toprule
        \multirow{1}{*}{Method} & Fusion Type &  F1-Score (\%) \\
        \hline
        \multirow{2}{*}{NSA} & SUM & 59.61\\
        & CONCAT & 45.10\\
        \hline
        \multirow{2}{*}{USA} & SUM & 60.78\\
        & CONCAT & 69.85\\
        \hline
        \multirow{1}{*}{{KEYLESS \cite{keyless}}} (2018) & CONCAT & 74.40\\
        \hline
        \multirow{1}{*}{{Best of UCSD-MIT\cite{mit_ucsd}}} (2019) & Early Fusion & 59.0 \\
        \hline
        \multirow{2}{*}{\textbf{\pa}} & \textbf{MAT-SUM} & \textbf{81.52}\\
        & MAT-CONCAT & 76.86\\
        \bottomrule
    \end{tabular}
     \vspace{-0.2in}
\end{table}

\begin{figure*}[!t]
  \centering
  \begin{tabular}{*{3}{c}}
    \includegraphics[width=.65\columnwidth]{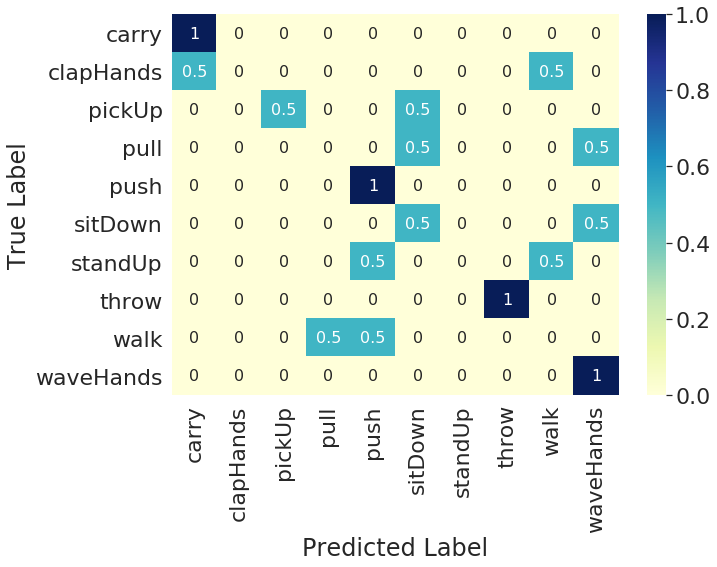} &
    \includegraphics[width=.65\columnwidth]{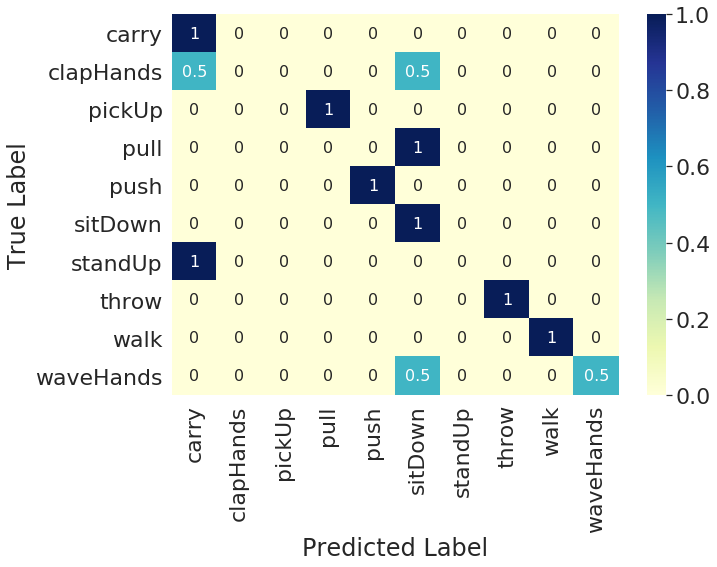} &
    \includegraphics[width=.65\columnwidth]{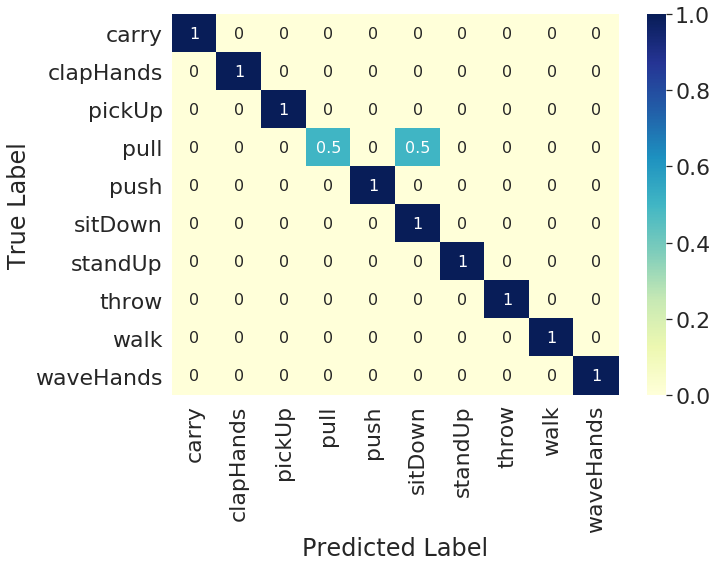} \\
    \footnotesize{(a) Without attention} & \footnotesize{(b) Unimodal attention} & \footnotesize{(c) Unimodal and multimodal attention} \\
  \end{tabular}
  \caption{Comparative impact of multimodal and unimodal attention in {\pa } for different activities on UT-Kinect dataset.}
  \label{fig:com_cm}
\end{figure*}

\begin{figure*}[!t]
  \centering
  \begin{tabular}{*{3}{c}}
    \includegraphics[width=.65\columnwidth]{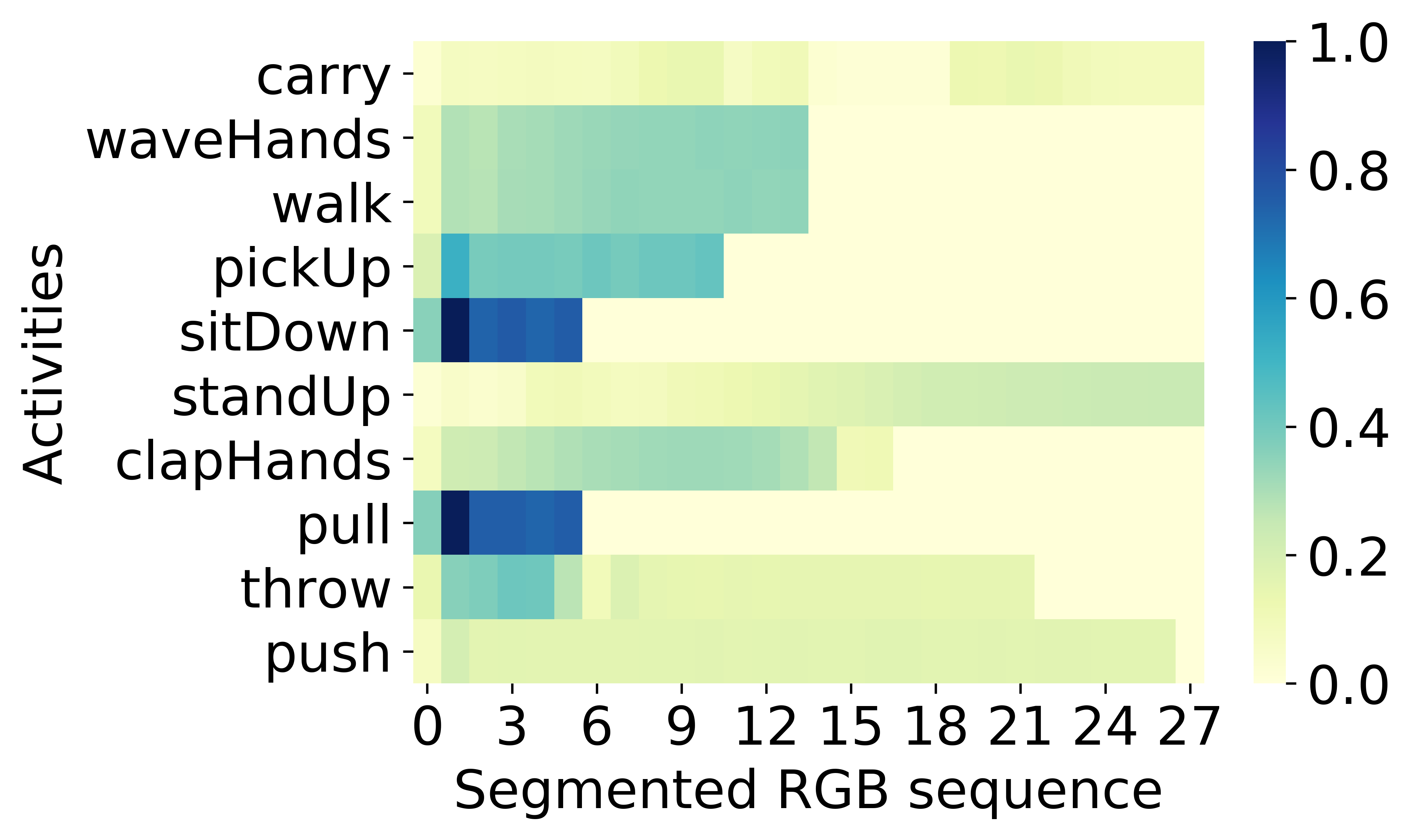} &
    \includegraphics[width=.65\columnwidth]{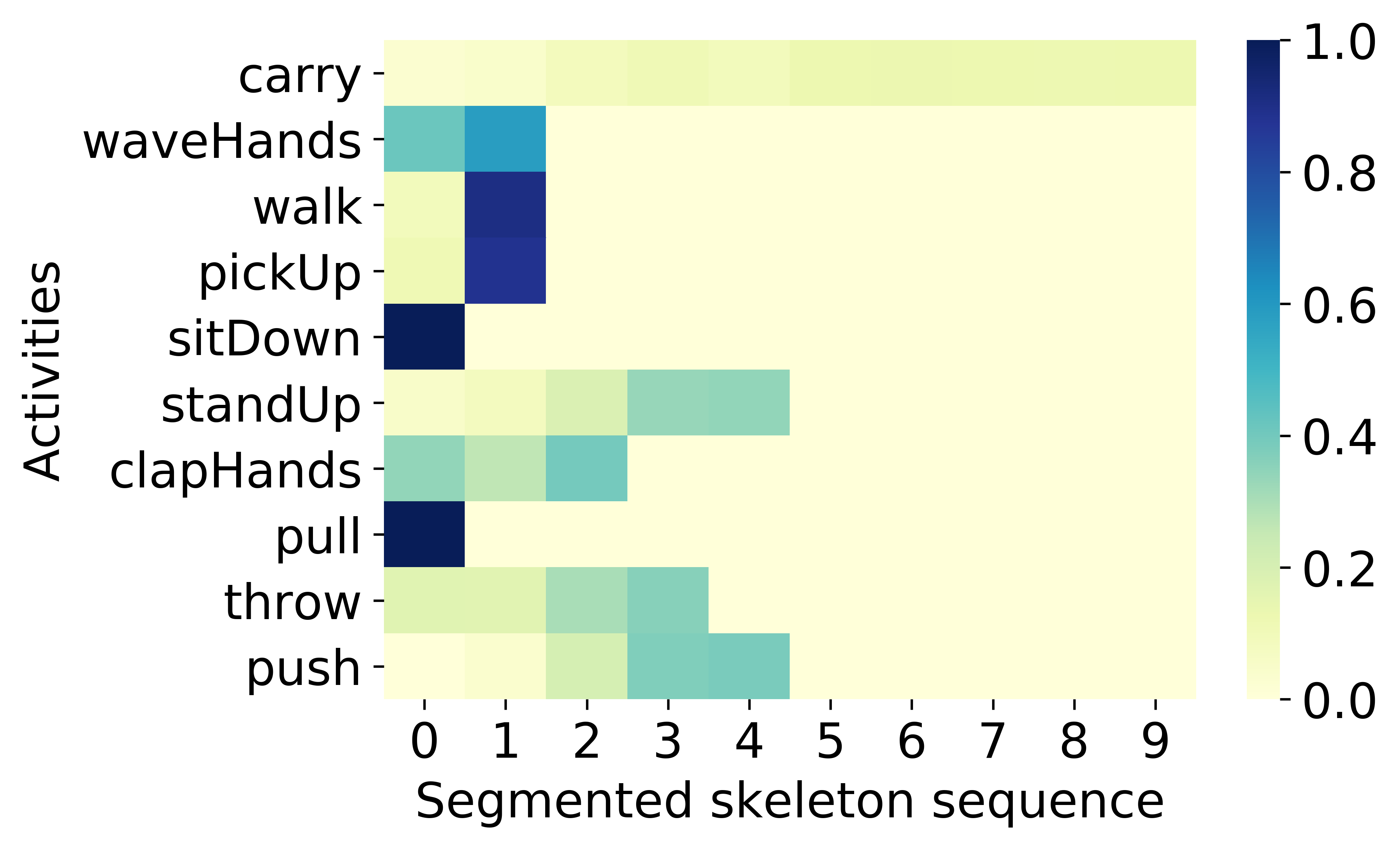} & \includegraphics[width=.65\columnwidth]{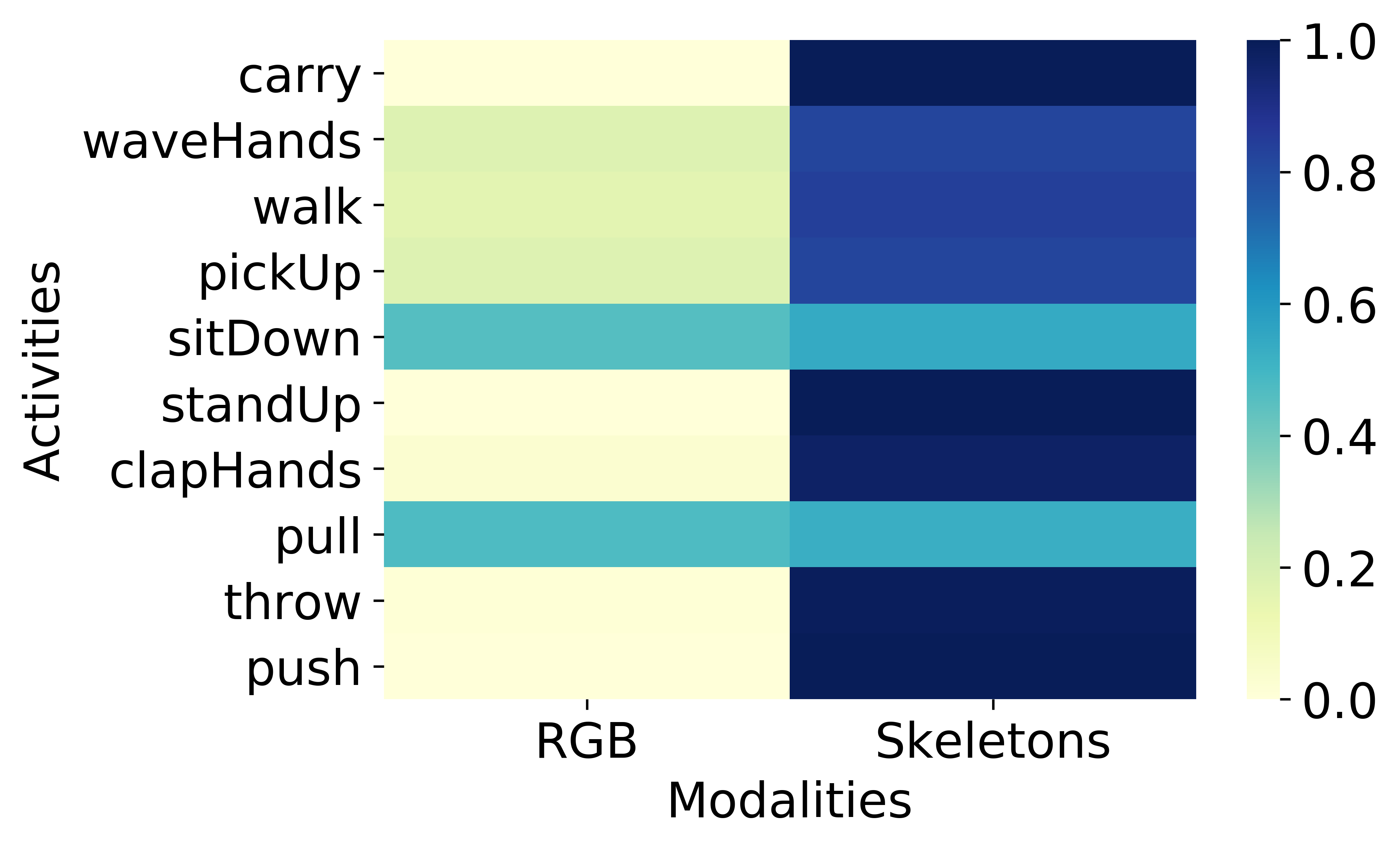} \\
    \footnotesize{(a) RGB sequence embedding attention} & \footnotesize{(b) Skeleton sequence embedding attention} & \footnotesize{(c) Multimodal fusion attention} \\
  \end{tabular}
  \caption{Multimodal and unimodal attention visualization for different activities on UT-Kinect Dataset.}
  \label{fig:vis_attention}
\end{figure*}

{\textbf{Discussion:}} {\pa } outperformed all other evaluated baselines across all datasets and metrics tested. The results on the UTD-MHAD dataset suggest that {\pa } outperformed all the state-of-the-art multimodal HAR methods. These methods didn't leverage the attention-based approaches to dynamically weighting the unimodal features to generate multimodal features. The results also suggest that, the other attention-based approaches, such as USA and Keyless \cite{keyless}, also showed better performance compared to the non-attention based approaches on UT-Kinect (Table~\ref{tab:com_on_ut_kinect}) and UCSD-MIT (Table~\ref{tab:com_on_ut_kinect}) datasets. The overall results support that our proposed approach is robust in finding appropriate multimodal features, hence it has achieved the highest HAR~accuracies.

\par The results indicate that the MAT-CONCAT approach achieved higher accuracy on the UT-Kinect dataset; however, the MAT-SUM approach delivered higher accuracy on the UCSD-MIT dataset. One explanation behind this variation is that the modalities (skeleton, sEMG, and IMUs) in the UCSD-MIT dataset represent similar physical body features, thus summing up the feature vectors work well. However, as the UT-Kinect dataset modalities have different characteristics, the visual (RGB) and the physical body (skeleton) features, MAT-CONCAT works better than MAT-SUM.

\par Finally, the overall results suggest that {\pa} achieved the mean F-1 score of 81.52\% on the UCSD-MIT dataset, which is lower compared to the highest accuracy on other datasets (please note that the top-1 accuracies were presented for other datasets). The main reason behind this performance degradation in UCSD-MIT is that this dataset contains missing data, especially sEMG, and IMUs data are missing in many instances. However, in the presence of the missing information, {\pa } showed the best performance compared to all other approaches.

\subsection{Combined Impact of Unimodal and Multimodal Attention}
\par We evaluated the comparative importance of unimodal and multimodal attention mechanism (presented in Fig.~\ref{fig:com_cm}). We can observe that the incorporation of unimodal attention (Fig.~\ref{fig:com_cm}-b) can help to reduce the miss-classification error in comparison to the non-attention based feature learning method (Fig.~\ref{fig:com_cm}-a). This is because unimodal attention can able to extract the sparse salient spatio-temporal features. We also can observe an improved accuracy in activity classification when the multimodal attention based unimodal feature fusion approach was incorporated (Fig.~\ref{fig:com_cm}-c vs. a, b). The results indicate that {\pa } can reduce the number of miss-classification, especially in the cases of similar activities, such as sitDown and pickUp, which is depicted in the confusion matrix in Fig.~\ref{fig:com_cm}-c.

\subsection{Visualizing Impact of Multimodal Attention: \hattn}
\label{sec:attention_vis}

We visualize the attention map of the unimodal and multimodal feature encoders to gauge the impact of attention in local (unimodal) and global (multimodal) feature representation in Fig~\ref{fig:vis_attention}. We used the data of the eighth performer from the UT-Kinect dataset \cite{ut_kinect} as a sample data to produce the attention map for different activities, as shown in Fig.~\ref{fig:vis_attention}, where we observe that the unimodal attention is able to detect salient segments of RGB (Fig~\ref{fig:vis_attention}-a) and skeleton (Fig~\ref{fig:vis_attention}-b) modalities. For example, the unimodal attention method focuses on the beginning parts of the \textit{sitDown} and the \textit{pull} activities, as these activities have distinguishable actions in the beginning parts of the activity. On the other hand, the unimodal attention method needs to pay attention to the full sequence to differentiate the \textit{carry} and the \textit{push} activities, as a specific part of these activities are not more informative than the other parts.

\par Moreover, we evaluate the impact of {\hattn } by observing the multimodal attention map in Fig.~\ref{fig:vis_attention}-c, which represents the relative attention given to unimodal features. For example, the \textit{pickUp} and \textit{sitDown} may involve similar skeleton joints movements, and thus if we concentrate only on the skeleton data, it may be challenging to differentiate between these two activities. However, if we incorporate the complementary modalities, such as RGB and skeleton, it may be easier to differentiate between similar activities. Thus, {\hattn} pays equal attention to the RGB and skeleton data while recognizing the \textit{sitDown} activity, whereas solely pay attention to the skeleton data while identifying the \textit{pickUp} activity (Fig.~\ref{fig:vis_attention}-c).  

\section{Conclusion}
\label{sec:conclusion}


In this paper, we presented {\pa }, a novel multimodal human activity recognition algorithm, for collaborative robotic systems. 
{\pa } first extracts the spatio-temporal salient features from the unimodal data and then employs a novel multimodal attention mechanism for disentangling and fusing the unimodal features for activity recognition. The experimental results suggest that {\pa } outperformed all other evaluated baselines across all datasets and metrics tested for human activity recognition.

\par In the future, we plan to implement {\pa } on a robotic system to enable it to perform collaborative activities in close proximity with people in an industrial environment. We also plan to extend {\pa } so that it can appropriately learn the relationship among the data from the modalities to address the missing data problem.

\bibliographystyle{IEEEtran}

{\footnotesize
\bibliography{references}

\begin{thebibliography}{10}
\providecommand{\url}[1]{#1}
\csname url@rmstyle\endcsname
\providecommand{\newblock}{\relax}
\providecommand{\bibinfo}[2]{#2}
\providecommand\BIBentrySTDinterwordspacing{\spaceskip=0pt\relax}
\providecommand\BIBentryALTinterwordstretchfactor{4}
\providecommand\BIBentryALTinterwordspacing{\spaceskip=\fontdimen2\font plus
\BIBentryALTinterwordstretchfactor\fontdimen3\font minus
  \fontdimen4\font\relax}
\providecommand\BIBforeignlanguage[2]{{%
\expandafter\ifx\csname l@#1\endcsname\relax
\typeout{** WARNING: IEEEtran.bst: No hyphenation pattern has been}%
\typeout{** loaded for the language `#1'. Using the pattern for}%
\typeout{** the default language instead.}%
\else
\language=\csname l@#1\endcsname
\fi
#2}}

\bibitem{utd_mhad}
C.~{Chen}, R.~{Jafari}, and N.~{Kehtarnavaz}, ``Utd-mhad: A multimodal dataset
  for human action recognition utilizing a depth camera and a wearable inertial
  sensor,'' in \emph{2015 IEEE ICIP}, Sep. 2015, pp. 168--172.

\bibitem{ut_kinect}
L.~Xia, C.~Chen, and J.~Aggarwal, ``View invariant human action recognition
  using histograms of 3d joints,'' in \emph{CVPRW}.\hskip 1em plus 0.5em minus
  0.4em\relax IEEE, 2012, pp. 20--27.

\bibitem{mit_ucsd}
A.~Kubota, T.~Iqbal, J.~A. Shah, and L.~D. Riek, ``Activity recognition in
  manufacturing: The roles of motion capture and semg+ inertial wearables in
  detecting fine vs. gross motion,'' in \emph{2019 ICRA}.\hskip 1em plus 0.5em
  minus 0.4em\relax IEEE, 2019, pp. 6533--6539.

\bibitem{Riek2017HealthCare}
L.~Riek, ``Healthcare robotics,'' \emph{Communications of the ACM}, 2017.

\bibitem{iqbal2019human}
T.~Iqbal and L.~D. Riek, ``Human-robot teaming: Approaches from joint action
  and dynamical systems,'' \emph{Humanoid robotics: A reference}, pp.
  2293--2312, 2019.

\bibitem{Iqbal2016T-RO}
T.~Iqbal, S.~Rack, and L.~D. Riek, ``Movement coordination in human-robot
  teams: A dynamical systems approach,'' \emph{IEEE Transactions on Robotics},
  vol.~32, no.~4, pp. 909--919, 2016.

\bibitem{andi_iros}
A.~E. Frank, A.~Kubota, and L.~D. Riek, ``Wearable activity recognition for
  robust human-robot teaming in safety-critical environments via hybrid neural
  networks,'' in \emph{IEEE/RSJ IROS}, 2019, pp. 449--454.

\bibitem{fosapt}
T.~{Iqbal}, S.~{Li}, C.~{Fourie}, B.~{Hayes}, and J.~A. {Shah}, ``Fast online
  segmentation of activities from partial trajectories,'' in \emph{2019 ICRA},
  May 2019, pp. 5019--5025.

\bibitem{iqbal2017coordination}
T.~Iqbal and L.~D. Riek, ``Coordination dynamics in multihuman multirobot
  teams,'' \emph{IEEE RA-L}, vol.~2, no.~3, pp. 1712--1717, 2017.

\bibitem{tiqbal_joint_action}
T.~{Iqbal}, M.~J. {Gonzales}, and L.~D. {Riek}, ``Joint action perception to
  enable fluent human-robot teamwork,'' in \emph{2015 24th IEEE RO-MAN}, Aug
  2015, pp. 400--406.

\bibitem{Iqbal2015TAC}
T.~Iqbal and L.~D. Riek, ``{A {M}ethod for {A}utomatic {D}etection of
  {P}sychomotor {E}ntrainment},'' \emph{IEEE Transactions on Affective
  Computing}, vol.~7, no.~1, pp. 3--16, 2016.

\bibitem{new_sk_rep}
Q.~Ke, M.~Bennamoun, S.~An, F.~Sohel, and F.~Boussaid, ``A new representation
  of skeleton sequences for 3d action recognition,'' in \emph{CVPR}, 2017, pp.
  3288--3297.

\bibitem{st_graph_sk}
S.~Yan, Y.~Xiong, and D.~Lin, ``Spatial temporal graph convolutional networks
  for skeleton-based action recognition,'' in \emph{Thirty-second AAAI
  conference on artificial intelligence}, 2018.

\bibitem{space_time_sk_review}
F.~Han, B.~Reily, W.~Hoff, and H.~Zhang, ``Space-time representation of people
  based on 3d skeletal data: A review,'' \emph{Computer Vision and Image
  Understanding}, vol. 158, pp. 85--105, 2017.

\bibitem{iqbal2016tempo}
T.~Iqbal, M.~Moosaei, and L.~D. Riek, ``Tempo adaptation and anticipation
  methods for human-robot teams,'' in \emph{RSS, Planning HRI: Shared Autonomy
  Collab. Robot. Workshop}, 2016.

\bibitem{multimodal_survey}
T.~{Baltrušaitis}, C.~{Ahuja}, and L.~{Morency}, ``Multimodal machine
  learning: A survey and taxonomy,'' \emph{IEEE Transactions on Pattern
  Analysis and Machine Intelligence}, vol.~41, no.~2, pp. 423--443, 2019.

\bibitem{self_attention_iros19}
G.~{Liu}, J.~{Qian}, F.~{Wen}, X.~{Zhu}, R.~{Ying}, and P.~{Liu}, ``Action
  recognition based on 3d skeleton and rgb frame fusion,'' in \emph{2019
  IEEE/RSJ IROS}, Nov 2019, pp. 258--264.

\bibitem{keyless}
X.~Long, C.~Gan, G.~De~Melo, X.~Liu, Y.~Li, F.~Li, and S.~Wen, ``Multimodal
  keyless attention fusion for video classification,'' in \emph{Thirty-Second
  AAAI Conference on Artificial Intelligence}, 2018.

\bibitem{fusion_approaches}
S.~M\"{u}nzner, P.~Schmidt, A.~Reiss, M.~Hanselmann, R.~Stiefelhagen, and
  R.~D\"{u}richen, ``Cnn-based sensor fusion techniques for multimodal human
  activity recognition,'' in \emph{Proceedings of the 2017 ACM ISWC}, 2017, p.
  158–165.

\bibitem{Hasan_2019}
M.~K. Hasan, W.~Rahman, A.~Bagher~Zadeh, J.~Zhong, M.~I. Tanveer, L.-P.
  Morency, and M.~E. Hoque, ``Ur-funny: A multimodal language dataset for
  understanding humor,'' \emph{EMNLP-IJCNLP}, 2019.

\bibitem{posemap}
M.~Liu and J.~Yuan, ``Recognizing human actions as the evolution of pose
  estimation maps,'' in \emph{CVPR}, 2018, pp. 1159--1168.

\bibitem{sdd_iccv}
P.~Wang, S.~Wang, Z.~Gao, Y.~Hou, and W.~Li, ``Structured images for rgb-d
  action recognition,'' in \emph{CVPRW}, 2017, pp. 1005--1014.

\bibitem{mcrl}
T.~{Liu}, J.~{Kong}, and M.~{Jiang}, ``Rgb-d action recognition using
  multimodal correlative representation learning model,'' \emph{IEEE Sensors
  Journal}, vol.~19, no.~5, pp. 1862--1872, 2019.

\bibitem{sos}
Y.~Hou, Z.~Li, P.~Wang, and W.~Li, ``Skeleton optical spectra-based action
  recognition using convolutional neural networks,'' \emph{IEEE Transactions on
  Circuits and Systems for Video Technology}, vol.~28, no.~3, pp. 807--811,
  2016.

\bibitem{jdm_cnn}
C.~Li, Y.~Hou, P.~Wang, and W.~Li, ``Joint distance maps based action
  recognition with convolutional neural networks,'' \emph{IEEE Signal
  Processing Letters}, vol.~24, no.~5, pp. 624--628, 2017.

\bibitem{dcnn}
J.~{Imran} and P.~{Kumar}, ``Human action recognition using rgb-d sensor and
  deep convolutional neural networks,'' in \emph{ICACCI}, 2016.

\bibitem{dmm_mff}
M.~F. Bulbul, Y.~Jiang, and J.~Ma, ``Dmms-based multiple features fusion for
  human action recognition,'' \emph{IJMDEM}, vol.~6, no.~4, pp. 23--39, 2015.

\bibitem{hussein2013human}
M.~E. Hussein, M.~Torki, M.~A. Gowayyed, and M.~El-Saban, ``Human action
  recognition using a temporal hierarchy of covariance descriptors on 3d joint
  locations,'' in \emph{Twenty-Third AAAI}, 2013.

\bibitem{co_occurrence}
C.~Li, Q.~Zhong, D.~Xie, and S.~Pu, ``Co-occurrence feature learning from
  skeleton data for action recognition and detection with hierarchical
  aggregation,'' in \emph{IJCAI}, 2018, p. 786–792.

\bibitem{closer_look_sp}
D.~Tran, H.~Wang, L.~Torresani, J.~Ray, Y.~LeCun, and M.~Paluri, ``A closer
  look at spatiotemporal convolutions for action recognition,'' in \emph{CVPR},
  2018, pp. 6450--6459.

\bibitem{sp_temporal_relation}
B.~Zhou, A.~Andonian, A.~Oliva, and A.~Torralba, ``Temporal relational
  reasoning in videos,'' in \emph{ECCV}, 2018, pp. 803--818.

\bibitem{sp_3d_conv}
D.~Tran, L.~Bourdev, R.~Fergus, L.~Torresani, and M.~Paluri, ``Learning
  spatiotemporal features with 3d convolutional networks,'' in \emph{ICCV},
  2015, pp. 4489--4497.

\bibitem{slowfast}
C.~Feichtenhofer, H.~Fan, J.~Malik, and K.~He, ``Slowfast networks for video
  recognition,'' in \emph{2019}, 2019.

\bibitem{totty2017muscle}
M.~S. Totty and E.~Wade, ``Muscle activation and inertial motion data for
  noninvasive classification of activities of daily living,'' \emph{IEEE
  Transactions on Biomedical Engineering}, vol.~65, no.~5, pp. 1069--1076,
  2017.

\bibitem{sp_3d_conv_lstm}
X.~Wang, L.~Gao, J.~Song, and H.~Shen, ``Beyond frame-level cnn: saliency-aware
  3-d cnn with lstm for video action recognition,'' \emph{IEEE Signal
  Processing Letters}, vol.~24, no.~4, pp. 510--514, 2016.

\bibitem{Garcia_2018_ECCV}
N.~C. Garcia, P.~Morerio, and V.~Murino, ``Modality distillation with multiple
  stream networks for action recognition,'' in \emph{ECCV}, 2018.

\bibitem{joze2019mmtm}
H.~R.~V. Joze, A.~Shaban, M.~L. Iuzzolino, and K.~Koishida, ``{MMTM}:
  Multimodal transfer module for cnn fusion,'' in \emph{CVPR}, 2020.

\bibitem{two_stream_cnn}
K.~Simonyan and A.~Zisserman, ``Two-stream convolutional networks for action
  recognition in videos,'' in \emph{NeurIPS}, 2014, pp. 568--576.

\bibitem{sp_two_stream_residual}
C.~Feichtenhofer, A.~Pinz, and R.~P. Wildes, ``Spatiotemporal residual networks
  for video action recognition,'' in \emph{Proceedings of the 30th
  NeurIPS'16}.\hskip 1em plus 0.5em minus 0.4em\relax Red Hook, NY, USA: Curran
  Associates Inc., 2016, p. 3476–3484.

\bibitem{sp_mul_motion_gating}
------, ``Spatiotemporal multiplier networks for video action recognition,'' in
  \emph{CVPR}, 2017, pp. 4768--4777.

\bibitem{conv_fusion}
C.~Feichtenhofer, A.~Pinz, and A.~Zisserman, ``Convolutional two-stream network
  fusion for video action recognition,'' in \emph{CVPR}, 2016, pp. 1933--1941.

\bibitem{zhang2018fusing}
S.~Zhang, Y.~Yang, J.~Xiao, X.~Liu, Y.~Yang, D.~Xie, and Y.~Zhuang, ``Fusing
  geometric features for skeleton-based action recognition using multilayer
  lstm networks,'' \emph{IEEE Transactions on Multimedia}, vol.~20, no.~9, pp.
  2330--2343, 2018.

\bibitem{mfas}
J.-M. Perez-Rua, V.~Vielzeuf, S.~Pateux, M.~Baccouche, and F.~Jurie, ``Mfas:
  Multimodal fusion architecture search,'' in \emph{CVPR}, June 2019.

\bibitem{attention}
D.~Bahdanau, K.~Cho, and Y.~Bengio, ``Neural machine translation by jointly
  learning to align and translate,'' in \emph{ICLR}, 2015.

\bibitem{attention_effective_approach}
M.-T. Luong, H.~Pham, and C.~D. Manning, ``Effective approaches to
  attention-based neural machine translation,'' \emph{arXiv preprint
  arXiv:1508.04025}, 2015.

\bibitem{xu2015show}
K.~Xu, J.~Ba, R.~Kiros, K.~Cho, A.~Courville, R.~Salakhudinov, R.~Zemel, and
  Y.~Bengio, ``Show, attend and tell: Neural image caption generation with
  visual attention,'' in \emph{ICML}, 2015, pp. 2048--2057.

\bibitem{lu2017knowing}
J.~Lu, C.~Xiong, D.~Parikh, and R.~Socher, ``Knowing when to look: Adaptive
  attention via a visual sentinel for image captioning,'' in \emph{CVPR}, 2017,
  pp. 375--383.

\bibitem{mnih2014recurrent}
V.~Mnih, N.~Heess, A.~Graves, \emph{et~al.}, ``Recurrent models of visual
  attention,'' in \emph{NeurIPS}, 2014, pp. 2204--2212.

\bibitem{lu2019vilbert}
J.~Lu, D.~Batra, D.~Parikh, and S.~Lee, ``Vilbert: Pretraining task-agnostic
  visiolinguistic representations for vision-and-language tasks,'' in
  \emph{NeurIPS}, 2019.

\bibitem{gao2019multi}
P.~Gao, H.~You, Z.~Zhang, X.~Wang, and H.~Li, ``Multi-modality latent
  interaction network for visual question answering,'' in \emph{ICCV}, 2019.

\bibitem{transformer}
A.~Vaswani, N.~Shazeer, N.~Parmar, J.~Uszkoreit, L.~Jones, A.~N. Gomez,
  {\L}.~Kaiser, and I.~Polosukhin, ``{Attention is all you need},''
  \emph{NeurIPS}, vol. 2017-Decem, no. NeurIPS, pp. 5999--6009, 2017.

\bibitem{adamw}
I.~Loshchilov and F.~Hutter, ``Decoupled weight decay regularization,'' in
  \emph{ICLR}, 2019.

\bibitem{resnet}
K.~He, X.~Zhang, S.~Ren, and J.~Sun, ``Deep residual learning for image
  recognition,'' in \emph{CVPR}, 2016, pp. 770--778.

\bibitem{warm_restart}
I.~Loshchilov and F.~Hutter, ``Sgdr: Stochastic gradient descent with warm
  restarts,'' in \emph{ICLR}, 2017.

\bibitem{paszke2019pytorch}
A.~Paszke, S.~Gross, F.~Massa, A.~Lerer, J.~Bradbury, G.~Chanan, T.~Killeen,
  Z.~Lin, N.~Gimelshein, L.~Antiga, \emph{et~al.}, ``Pytorch: An imperative
  style, high-performance deep learning library,'' in \emph{NeurIPS}, 2019, pp.
  8024--8035.

\end{thebibliography}
}
\end{document}